# Harmony Search Algorithm for Curriculum-Based Course Timetabling Problem

Juliana Wahid
School of Computing
Universiti Utara Malaysia
Kedah, Malaysia
w.juliana@uum.edu.my

Naimah Mohd Hussin
Faculty of Computer Science and Mathematics
Universiti Teknologi MARA
Perlis, Malaysia
naimahmh@perlis.uitm.edu.my

*Abstract*— In this paper, harmony search algorithm is applied to curriculum-based course timetabling. The implementation, specifically the process of improvisation consists of memory consideration, random consideration and pitch adjustment. In memory consideration, the value of the course number for new solution was selected from all other course number located in the same column of the Harmony Memory. This research used the highest occurrence of the course number to be scheduled in a new harmony. The remaining courses that have not been scheduled by memory consideration will go through random consideration, i.e. will select any feasible location available to be scheduled in the new harmony solution. Each course scheduled out of memory consideration is examined as to whether it should be pitch adjusted with probability of eight procedures. However, the algorithm produced results that were not comparatively better than those previously known as best solution. With proper modification in terms of the approach in this algorithm would make the algorithm perform better on curriculum-based course timetabling.

I. INTRODUCTION

The university course timetabling is the administrative tasks that allocate the set of courses offered by the university to particular rooms and time slots. The allocating process needs to satisfy several constraints so that timetable can actually be carried out (feasible). There are two categories of constraints in timetabling such as hard and soft constraints [1]. Hard constraints or constraints which may not be broken consist of matters that need rigidly be fulfilled in the timetable such as courses must be allocated into different time slots in order to avoid students and staffs clashes. While soft constraints are matters reflecting the preferences and teaching/learning comfort expectations of the students and staffs which are not necessarily essential. Some examples of soft constraints are such as avoiding students having to attend three or more courses in successive time slots, and stopping students from having only one course in any day.

The university course timetabling as described in International Timetabling Competition website (www.cs.qub.ac.uk/itc2007/) fall into two versions. The first version is the Post Enrolment, which the timetable is constructed based on student enrolments i.e. after students have selected which lectures they wish to attend. The second version is the curriculum-based course timetabling (CB-CTT). This version will construct the timetable according to curricula published by the university.

University course timetabling problem (UCTP) that fall into group of non deterministic polynomial (NP) problem can be formulated as a combinatorial optimization problem (COP) [2-3], in which the larger the number of lectures to be scheduled and the diversity of constraints need to be considered, the harder the problem to be solved.

The works towards solving course timetabling problems have started in year 1960's, whereby the problem is modeled as a graph coloring approach [4]. Since then, various methods and approaches have been explored and proposed to solve the problems. In year 1970's, Barham and Westwood [5] have attempted to use heuristic methods. Moving forward to year 1980's, Mulvey [6] solved the university course-scheduling problem using network flows while Werra and Hertz [7] have tested tabu search methods. As in 1990's, the works in this problems are more unleashed where variety of methods are proposed. Yu [8] used artificial neural networks to solve the timetabling problem. In the other hand, Downsland [9] and Dige et al. [10] solved the timetabling problem by simulated annealing, while Deris et al. [11] exhibited constraint-based reasoning method in a case study of university timetabling problem. The active timetable research in this era also was contributed by the First International Conference on the Practice and Theory of Automated Timetabling (PATAT) held in 1995. In this era also, the evolutionary algorithms [12] in timetabling problem are proposed. In the new millennium era, the works continue. Constraint based programming [13], hyper heuristic methods [14-15], and metaheuristic methods [16] has been introduced and become an interesting approach for solving timetabling problems.

Even though enormous effort expended over the last forty years to discover efficient methods and approaches for solving timetabling problems, these problems are nevertheless still the focus of intense research [17]. For course timetabling problems, the solutions algorithms are usually problem-specific; that is, no general solution can solve different problem instances. Even though the problem has been widely researched, there is still no definitive solution approach that is able to provide excellent solutions across the broad spectrum of problem instances. This is parallel with the No Free Lunch theorem [18] that says if we cannot make any prior assumptions about the optimization problem we are trying to solve, no strategy can be expected to perform better than any other. Indeed, Gomes & Williams [19] stated that NP-hard problems are inflexible, which means that there does not exist an efficient algorithm that is ensured to find an optimal solution for this kind of problems.

This study has interest on examining specific algorithm i.e. harmony search algorithm (HSA). There are several components that can be improved in finding solution to university course timetabling problem (UCTP). HSA was successfully adapted to UCTP by Al-Betar, Khader, & Abdul Gani [20]. In their work, Al-Betar, et al. [20] described the adaptation of HSA for UCTP using five steps of HSA such as 1: Initialize HSA parameters and UCTP parameters, 2: Initialize Harmony Memory (HM) with random feasible timetables based on Harmony Memory Size (HMS) Parameter, 3: Improvise new harmony solution, 4: Update harmony memory, and 5: Stop Criteria. The HSA can find near optimal solutions for UCTP and produces better results than several others in the previous literature.

In their later work [21], the modified harmony search algorithm (MHSA) was produced to enhance the basic HSA for UCTP whereby the basic memory consideration of basic HSA is changed to modified memory consideration, and pitch adjustment with random walk as an acceptance rule of basic HSA is changed to pitch adjustment with side walk and first improvement acceptance rule.

In the same year, works on HSA for UCTP focused on pitch adjustment operator produced whereby eight pitch adjustment procedures were proposed to enhance the solution quality of the UCTP [22].

A year before the producing of MHSA, the hybridization of HSA with hill climbing optimiser (HCO) and particle swarm optimization (PSO) was produced by Al-Betar & Khader [23]. The role of HCO in the hybridization is as new operator (located at step 3) to improve the quality of new harmony vector relative to the number of improvisations (NI) in each run with probability of hill climbing rate (HCR). The PSO was also located at step 3, used to modify the memory consideration operator to imitate the best harmony among the HM vector to construct new harmony.

As according to Al-Betar & Khader [23], the HSA exploits the advantages of population-based methods by identifying the potential regions in the search space using the memory consideration and random consideration operator. It also utilizes the advantages of local search-based methods by fine tuning the search space region using the pitch adjustment operators. So far in the literature, the works on HSA for UCTP carried out mostly focussed on the first version of UCTP, i.e. the Post Enrolment. None so far focussed on the curriculum-based course timetabling.

From this point of view, this study will explore the implementation of harmony search algorithm (HSA) into CB-CTT.

This paper is organized as follow: section 2 describes literature review on the curriculum-based course university timetabling problem and the details of the harmony search algorithm, section 3 presents the methodology used in the paper followed by section 4 that explains the details of implementation of harmony search algorithm for CB-CTT. Section 5 discusses the experimental results and compares them with the best known solution in the literature. The possible future approach for improvement also discussed in this section. Section 6 shows some conclusions.

II. LITERATURE REVIEW

A. *Curriculum-based course timetabling (CB-CTT)*

The curriculum-based course timetabling (CB-CTT) problem is an alternative of an university timetabling problem which creating a weekly timetable by allocating lectures for several university courses to a certain number of rooms and time periods based on curricula. The available constraints for this variant of university timetabling problem are solely based on the definition of the curricula [24]. Opposite to post-enrolment based course timetabling problems in which students specifically register for courses they wish to attend.

The curriculum-based course timetabling problem has been selected as one of the competition tracks of the International Timetabling Competition 2007 (ITC2007). This study used problem definition presented in Bonutti, De Cesco, Di Gaspero, & Schaerf [25] which extended from technical description in the ITC2007 web site and the corresponding technical report from Di Gaspero et al. [26]. The CB-CTT problem definition consists of the following basic entities:

**Days** – number of *teaching days* in the week (typically 5 or 6).

**Timeslots** - Each day is split into a fixed number of *timeslots*, which is equal for all days.

**Periods** - a pair composed of a day and a timeslot. The total number of scheduling periods is the product of the days times the day timeslots.

**Courses and Teachers** - Each course consists of a fixed *number of lectures* to be scheduled in distinct periods, it is attended by a given *number of students*, and is taught by a *teacher*. For each course there is a minimum number of days that the lectures of the course should be spread in, moreover there are some periods in which the course cannot be scheduled.

**Rooms** - Each *room* has a *capacity*, expressed in terms of number of available seats, and a *location* expressed as an integer value representing a separate building. Some rooms may not be suitable for some courses (because they miss some equipment).

**Curricula** - A *curriculum* is a group of courses such that any pair of courses in the group have students in common. Based on curricula, we have the *conflicts* between courses and other soft constraints.

The CB-CTT problem consists in scheduling lectures of a set of courses into a weekly timetable, where each lecture of a course must be assigned a period and a room in accordance with a given set of constraints. A feasible timetable is one in which all lectures have been scheduled at a timeslot and a room, so that the hard constraints such as follows are satisfied:

**H1 - Lectures**: Each lecture of a course must be scheduled in a distinct period and a room.

**H2 - Room occupancy**: Any two lectures cannot be assigned in the same period and the same room.

**H3 - Conflicts**: Lectures of courses in the same curriculum or taught by the same teacher cannot be scheduled in the same period, i.e., no period can have an overlapping of students nor teachers.

**H4 - Availability**: If the teacher of a course is not available at a given period, then no lectures of the course can be assigned to that period.

In addition, a feasible timetable satisfying the four hard constraints incurs a penalty cost for the violations of the four soft constraints such as follows:

**S1 - Room capacity**: For each lecture, the number of students attending the course should not be greater than the capacity of the room hosting the lecture.

**S2 - Room stability**: All lectures of a course should be scheduled in the same room. If this is impossible, the number of occupied rooms should be as few as possible.

**S3 - Minimum working days**: The lectures of a course should be spread into the given minimum number of days.

**S4 - Curriculum compactness**: For a given curriculum, a violation is counted if there is one lecture not adjacent to any other lecture belonging to the same curriculum within the same day.

Then, the objective of the CB-CTT problem is to minimize the number of soft constraint violations in a feasible solution.

*B. Harmony search algorithm (HSA)*

The harmony search algorithm (HSA) is a new metaheuristic algorithm that impersonates the musical improvisation process in which a group of musicians improvise their instruments' pitch by searching for a perfect state of harmony according to audio-aesthetic standard [27]. The equivalent terms between musical scenario and optimization problems that obtained from Al-Betar & Khader [21] are as the following:

- Improvisation ↔ Generation or construction
- Harmony ↔ Solution vector
- Musician ↔ Decision variable
- Pitch ↔ Value
- Pitch range ↔ Value range
- Audio-aesthetic standard ↔ Objective function
- Practice ↔ Iteration
- Pleasing harmony ↔ (Near-) optimal solution

The HSA consists of several steps such as 1) problem formulation, 2) algorithm parameter setting, 3) random tuning for memory initialization, 4) harmony improvisation (memory consideration, random consideration, and pitch adjustment), 5) memory update, 6) performing termination, and 7) cadenza [28]. Figure 1 shows the pseudo-code of the HSA with five main steps in which STEP1 consists of problem formulation and algorithm parameter setting, STEP2 consists of random tuning for memory initialization, STEP3 consists of harmony improvisation (memory consideration, random consideration, and pitch adjustment), STEP4 consists of memory update and STEP5 consists of performing termination and cadenza. The details of each step will further discussed in section 4.

```
1. STEP1. Initialize the problem and HSA parameters
     Input data. The data instance of the optimization problem and the HSA parameters
                 (HMCR, PAR, NI, HMS)
2. STEP2. Initialize the harmony memory
     Construct the vectors of the harmony memory, HM = {x¹, x², ..., x^HMS}
     Recognize the worst vector in HM, x^worst ∈ {x¹, x², ..., x^HMS}
3. STEP3. Improvise a new harmony
     x' = φ // new harmony vector
     for i = 1, ..., N do     // N is the number of decision variables.
       if (U(0, 1) ≤ HMCR) then    // U is a uniform random number generator.
       begin
         x'_i ∈ {x¹_i, x²_i, ..., x^HMS_i}   {* memory consideration *}
         if (U(0, 1) ≤ PAR) then
           x'_i = v_{i,k±m}     // x'_i = v_{i,k}    {*pitch adjustment *}
         end
       else
         x'_i ∈ X_i     {* random consideration *}
       end if
     end for
4. STEP4. Update the harmony memory (HM)
     if (f(x') < f(x^worst)) then
       Include x' to the HM.
       Exclude x^worst from HM.
5. STEP5. Check the stop criterion
     while (not termination criterion is specified by NI)
       Repeat STEP3 and STEP4
```

Fig. 1. Pseudo Code of Harmony Search Algorithm Source: [21]

### III. METHODOLOGY

This section list the steps used to achieve the objective in this study such as follows:

*A. Initial Phase*

This phase concentrates on studying and understanding the HSA and CB-CTT from literature that can be found in section 2.

### B. Preprocessing Phase

This phase focuses on understanding the original data file and transferring them into related matrixes, such as conflict matrix that represents a conflict between courses, where Lectures of courses in the same curriculum or taught by the same teacher must be all scheduled in different periods.

### C. Construction Phase

This phase focuses on finding feasible initial solutions that satisfy all hard constraints related to each data instances.

### D. Improvement Phase

This phase aims to improve solutions generated from the construction phase, where the HSA are employed.
The details of step ii, iii and iv are covered in the next section.

### E. Analysis Phase

This phase evaluates the quality of the solutions obtained from the improvement phase against the other methods in the literature which can be seen in section 5.

## IV. THE HARMONY SEARCH ALGORITHM FOR CB-CTT

This section describes details of steps as shown earlier in Figure 1 in implementing HSA into CB-CTT as follows:

### A. CB-CTT formulation and HSA parameter setting

The problem formulation of CB-CTT is based on the data instances provided by ITC2007 available on http://tabu.diegm.uniud.it/ctt. The details of the data instances are describes in Table 1.

TABLE 1. Description of CB-CTT Instances Source:[25]

| Instance | No. of courses | No. of lectures | No. of rooms | No. of periods per day | No. of Days | No. of Curriculums |
|---|---|---|---|---|---|---|
| comp01 | 30 | 160 | 6 | 6 | 5 | 14 |
| comp02 | 82 | 283 | 16 | 5 | 5 | 70 |
| comp03 | 72 | 251 | 16 | 5 | 5 | 68 |
| comp04 | 79 | 286 | 18 | 5 | 5 | 57 |
| comp05 | 54 | 152 | 9 | 6 | 6 | 139 |
| comp06 | 108 | 361 | 18 | 5 | 5 | 70 |
| comp07 | 131 | 434 | 20 | 5 | 5 | 77 |
| comp08 | 86 | 324 | 18 | 5 | 5 | 61 |
| comp09 | 76 | 279 | 18 | 5 | 5 | 75 |
| comp10 | 115 | 370 | 18 | 5 | 5 | 67 |
| comp11 | 30 | 162 | 5 | 9 | 5 | 13 |
| comp12 | 88 | 218 | 11 | 6 | 6 | 150 |
| comp13 | 82 | 308 | 19 | 5 | 5 | 66 |
| comp14 | 85 | 275 | 17 | 5 | 5 | 60 |
| comp15 | 72 | 251 | 16 | 5 | 5 | 68 |
| comp16 | 108 | 366 | 20 | 5 | 5 | 71 |
| comp17 | 99 | 339 | 17 | 5 | 5 | 70 |
| comp18 | 47 | 138 | 9 | 6 | 6 | 52 |
| comp19 | 74 | 277 | 16 | 5 | 5 | 66 |
| comp20 | 121 | 390 | 19 | 5 | 5 | 78 |
| comp21 | 94 | 327 | 18 | 5 | 5 | 78 |

The problem formulation of CB-CTT was to minimize the cost of soft constraints such as room capacity, room stability, minimum working days and curriculum compactness. The timetable solution for CB-CTT in HSA is represented by a vector of courses $x = (x1, x2, \ldots, xN)$, each lecture for each course must be scheduled in a feasible location. For example, in the first data problem instances established by ITC2007, the number of courses $C = 30$ which have number of lectures that sum up to total lectures $L = 160$, the number of rooms $R = 6$, the number of period per day $P = 6$ and the number of days $D = 5$. The possible timeslots of each lectures that represent by courses is within the range between 0 to 179 ($R \times P \times D = 6 \times 6 \times 5 = 180$). As shown in Figure 2, first course denotes by 0 located at first room in period 0, 1, 3 and 4 in Day 0, and period 0 and 2 in Day 1. These locations equivalent to timeslot 0, 1, 3, 4, 6, and 8. Solution with value -1 shows that the timeslot is empty.

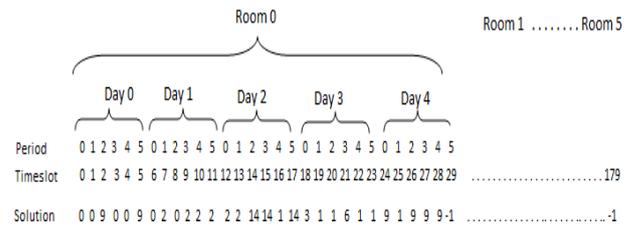

Fig. 2. Structure of Timeslots against Solution

As for the preprocessing phase, the following data structures are used to build a CB-CTT solution:

- Course Correlation matrix: is a matrix with size C x C where the element > 0 if courses in the same curriculum and if more then one course tought by same teacher or 0 otherwise.

- Course period matrix: is a binary matrix with size C x P where the element contains either 1 if and only if course C not available in period P or 0 otherwise.

- Course room matrix: is a binary matrix with size C x R where the element contains either 1 if and only if course C and room R is compatible with both aspects of size and features or 0 otherwise.

- Courseperiodavail matrix C x (P xR) contains either 1 if and only if course C available in period P in room R or 0 otherwise.

The following parameters of the HSA that identified by Geem [28] which required to solve the optimization problem are also specified in this step:

- Harmony memory size (HMS) - number of solution vectors concurrently handled.
- Harmony memory considering rate (HMCR) – ($0 \leq$ HMCR $\leq 1$) where HSA picks one value randomly from memory.
- Pitch adjusting rate (PAR) - ($0 \leq$ PAR $\leq 1$) is the rate where HSA fine-tunes the value which was originally picked from memory.

- Maximum improvisation (MI) - number of iterations or stopping criterion.

The harmony memory (HM) is a memory location where all the solution vectors (sets of decision variables) and corresponding objective function values are stored. The function values are used to evaluate the quality of solution vectors. In CB-CTT, the HM consists of sets of timetable solution and corresponding of costs of soft constraints i.e. room capacity, room stability, minimum working days and curriculum compactness. As surveyed by Geem [29], HMS of value range from 30 to 100 was frequently used in literature.

The HSA parameter setting used in this study are as follows:

- Harmony memory size (HMS) - 50
- Harmony memory considering rate (HMCR) – 0.9
- Pitch adjusting rate (PAR) – 1.0, with multi pitch with eight procedures [22] such as follows:
  - Move-timeslot: $0 < U(0,1) \leq 0.10 \times PAR$
  - Swap-timeslot: $0.10 \times PAR < U(0,1) \leq 0.20 \times PAR$
  - move-location: $0.20 \times PAR < U(0,1) \leq 0.30 \times PAR$
  - swap-location: $0.30 \times PAR < U(0,1) \leq 0.40 \times PAR$
  - exchange-location: $0.40 \times PAR < U(0,1) \leq 0.50 \times PAR$
  - swap-distinct-timeslots: $0.50 \times PAR < U(0,1) \leq 0.60 \times PAR$
  - move-room: $0.60 \times PAR < U(0,1) \leq 0.70 \times PAR$
  - Swap-room: $0.70 \times PAR < U(0,1) \leq 0.80 \times PAR$
  - do nothing: $0.80 \times PAR < U(0,1) \leq 1$
- Maximum improvisation (MI) – first phase: 10, Second phase: 50

### B. Random tuning for memory initialization

In this step, the harmony memory (HM) matrix is filled with as many randomly generated solution vectors as the HMS. In the CB-CTT model, the randomly generated solution vectors undergo a validation process to verify that they feasible i.e. not violated the hard constraints (lectures, conflicts, availability and room occupation).

This step constructs maximum feasible timetabling solutions for CB-CTT as determined by HMS. HM is filled with these solutions and sorted according to the minimum total cost of soft constraints. The technique used to generate random HM solutions is assigning the courses by using the combination of graph heuristic that produced maximum numbers of solutions proposed by Juliana & Naimah [30] which combines saturation degree with largest degree. In this technique, the course with the least available period and largest number soft conflicting students is scheduled first.

### C. Harmony improvisation (memory consideration, random consideration, and pitch adjustment)

#### 1) Memory consideration

The memory consideration selects feasible locations of the courses to be scheduled in the new harmony solution, $x\_ = (x1, x2, \ldots, xN)$, from the solutions stored in HM with the probability of HMCR. In memory consideration, the value of the course number for new solution was selected from all other course number located in the same column of the HM. This research used the highest occurrence of the course number to be scheduled in a new harmony with probability of HMCR. In certain situations, if there are same numbers of highest occurrence for any course number, the least available period and highest conflicts for that course were used to choose between the course numbers. If the operators cannot find a feasible timetable (this happens in some medium and large data instances), the algorithm initiates a neighborhood search to find feasible timetable, such as find an empty timeslot for unassigned course number and swapping between course numbers.

#### 2) Random consideration

The remaining courses that have not been scheduled by memory consideration will select any feasible location available to be scheduled in the new harmony solution with probability (1-HMCR).

#### 3) Pitch adjustment

In CB-CTT the pitch adjustment operator works similar to neighborhood structures in local search-based methods into eight procedures [22] as follows:

a) The pitch adjustment *move-timeslot*. An event that meets the probability of 10%×PAR is randomly *moved* to any feasible timeslot where the room is not changed.

b) The pitch adjustment *Swap-timeslot*. An event that meets the probability between 10%×PAR and 20%×PAR is swapped with the timeslot of another event, while the rooms of both events are not changed.

c) The pitch adjustment *move-location*. An event that meets the probability between 20%×PAR and 30%×PAR is randomly *moved* to any free feasible location in the new harmony solution.

d) The pitch adjustment *swap-location*. An event that meets the probability between 30%×PAR and 40%×PAR is randomly *swapped* with another event while the feasibility is maintained.

e) The pitch adjustment *exchange-location*. An event that meets the probability between 40%×PAR and 50%×PAR is randomly *exchanged* with another two events while the feasibility is maintained.

f) The pitch adjustment *swap-distinct-timeslots*. An event that meets the probability between 50%×PAR and 60%×PAR is adjusted as follows: (1) select all the events that have the same timeslot as first event. (2) select a timeslot in random. (3) simply swap all the events in timeslot with all the events in other timeslot without changing the rooms.

g) The pitch adjustment *move-room*. An event that meets the probability between 60%×PAR and 70%×PAR is *moved* to any free feasible room while the timeslot is not changed.

h) The pitch adjustment *Swap-room*. An event that meets the probability between 70%×PAR and 80%×PAR swaps its room with a room of another event in the same timeslot while reserving the feasibility.

Each course scheduled out of memory consideration is examined as to whether it should be pitch adjusted with probability by the above eight procedures.

*4) Repair process*

Due to large size of data instances, some courses may not able to find feasible locations in the new harmony. This extra step needs to be carried out to ensure all courses are scheduled in the new harmony solutions. The repair process used here is based on a one-level backtracking process by Al-Betar & Khader [21] with some modification such as follows:
a) Select an unscheduled course.
b) Find all feasible locations for the unscheduled course which is currently occupied by other courses in the new harmony solution.
c) Delete the course that held the feasible location and add it to the unscheduled list.
d) Schedule the unscheduled course to new harmony solution in the feasible location and remove it from the unscheduled list.

The above steps of repair process are carried out in iterative manner, however if the predefined iterations cannot find a complete feasible timetable, the improvisation process of new harmony is restarted with a new random seed and discarded the current new solution.

*D. Memory update*

If the new harmony vector, $x\_ = (x1, x2, . . . , xN)$, is better than the worst harmony stored in HM in terms of the objective function value, the new harmony vector is included to the HM, and the worst harmony vector is excluded from the HM.

*E. Performing termination*

Steps 4.3 and 4.4 of HSA are repeated until the number of maximum improvisation (MI) is met. For the purpose of this study two cycles of iterations, i.e. 10 and 50, were used to observe whether the solutions are improving or not.

## V. RESULTS AND DISCUSSIONS

In this section, the performance of HSA for solving the CB-CTT problem was evaluated using 21 data instances generated in ITC-2007. The proposed method is coded using C++ in Microsoft Visual 2008 under Windows Vista on an Intel Machine with Core TM and a 2.16GHz processor and 1GB RAM. Table 2 shows the experimental results of HSA algorithm for 10 iterations and 50 iterations together with the best known results solution. As there are several instances that not improving from 10 iterations to 50 iterations such as comp05, comp07, and comp12, the higher iterations e.g. 100, 150, and so on were not carried out yet to enhance the proposed method. The time complexity of each iteration also not yet described as this paper shows only low level number of iterations.

So far, the result demonstrates that HSA could be tailored to solve CB-CTT problems. However, our approach is still not able to achieve results comparable with the best known solution. An improvement (which is currently ongoing, with optimism) is required, in order to enhance the performance of the proposed method to produce desired outcomes better than those currently in use. There several issues that needed to be considered in enhancing the approach such as in the memory consideration, perhaps the used of smallest occurrence for course number can be applied. Another issue could be the process of repairing after improvisation process. A method to ensure all new solutions after improvisation process is feasible without to restart with a new random seed also be needed.

The method used in this study provide alternative way of solving curriculum-based course timetabling, as this type of data instance not yet focused by other researcher that involved in this HSA.

TABLE 2. Experimental Result of HSA for CB-CTT

| Problem instances | Initial Solution Penalty | HSA (10 iterations) Penalty | HSA (50 iterations) Penalty | Best Known Solution[*] (Until 24/5/2012) | |
|---|---|---|---|---|---|
| | | | | Penalty | Method Used |
| Comp01 | 323 | 323 | 322 | 5 | Tabu Search |
| Comp02 | 747 | 747 | 732 | 24 | SAT-based |
| Comp03 | 715 | 701 | 665 | 66 | Local Search |
| Comp04 | 692 | 617 | 577 | 35 | Local Search |
| Comp05 | 1297 | 1297 | 1297 | 290 | SA |
| Comp06 | 982 | 934 | 879 | 27 | SAT-based |
| Comp07 | 1063 | 930 | 930 | 6 | SAT-based |
| Comp08 | 788 | 663 | 645 | 37 | Other |
| Comp09 | 849 | 730 | 685 | 96 | Tabu Search |
| Comp10 | 920 | 868 | 816 | 4 | SAT-based |
| Comp11 | 215 | 208 | 179 | 0 | Tabu Search |
| Comp12 | 1542 | 1398 | 1398 | 300 | SA |
| Comp13 | 818 | 731 | 694 | 59 | Tabu Search |
| Comp14 | 720 | 707 | 702 | 51 | Mathematical Programming |
| Comp15 | 715 | 701 | 665 | 66 | Tabu Search |
| Comp16 | 965 | 831 | 827 | 18 | SAT-based |
| Comp17 | 910 | 859 | 830 | 56 | SAT-based |
| Comp18 | 627 | 534 | 510 | 62 | Hybrid Methods |
| Comp19 | 668 | 666 | 608 | 57 | Local Search |
| Comp20 | 1036 | 965 | 950 | 4 | SAT-based |
| Comp21 | 906 | 839 | 835 | 75 | Mathematical Programming |

## VI. CONCLUSIONS

This paper presented the harmony search algorithm for solving the CB-CTT problems. Results from the experiments have shown that the algorithm is capable of solving timetabling problems. Although the results produced by the algorithm in this study are presently not comparatively better than those already reported in the literature, there are rooms for improvement. Further works need to be considered to improve the HSA performance.

ACKNOWLEDGMENT

This study was funded by Ministry of Higher Education of Malaysia.